\documentclass{article} 
\usepackage{nips15submit_e,times}
\usepackage{hyperref}
\usepackage{url}

\usepackage{graphicx}

\usepackage{color}
\usepackage{xcolor}
\usepackage{subfig}
\usepackage{hhline}
\usepackage{caption}
\usepackage{multirow}
\title{Joint Line Segmentation and Transcription for End-to-End Handwritten Paragraph Recognition}

\author{
Th\'eodore Bluche \\
A2iA SAS\\
39 rue de la Bienfaisance\\
75008 Paris \\
\texttt{tb@a2ia.com} \\
}

\nipsfinalcopy 

\begin{document}

\maketitle

\newcommand{\ie}{\textit{i.e.} }
\newcommand{\eg}{\textit{e.g.} }
\newcommand{\x}{$\mathbf{x}$}
\newcommand{\p}{\textbf{}}
\newcommand{\bi}[1]{\textit{\textbf{#1}}}
\newcommand{\ijt}[3]{#1_{#2}^{(#3)}}

\newcommand{\fig}[1]{Figure~\ref{fig:#1}}
\newcommand{\sect}[1]{Section~\ref{sec:#1}}
\newcommand{\tab}[1]{Table~\ref{tab:#1}}
\newcommand{\eqn}[1]{Eqn.~\ref{eqn:#1}}

\begin{abstract}
Offline handwriting recognition systems require cropped text line images for 
both training and recognition. On the one hand, the annotation of position and transcript at 
line level is costly to obtain. On the other hand, automatic line segmentation 
algorithms are prone to errors, compromising the subsequent recognition. 
In this paper, we propose a modification of the popular and efficient 
multi-dimensional long short-term memory recurrent neural networks (MDLSTM-RNNs)
to enable end-to-end processing of handwritten paragraphs. More particularly,
we replace the collapse layer transforming the two-dimensional representation into
a sequence of predictions by a recurrent version which can recognize one line at a time. 
In the proposed model, a neural network performs a kind of implicit line segmentation by computing attention weights  
on the image representation. The experiments on paragraphs of Rimes and IAM database
yield results that are competitive with those of networks trained at line level,
and constitute a significant step towards end-to-end transcription of full documents.
\end{abstract}

\section{Introduction}

Offline handwriting recognition consists in recognizing a sequence of characters 
in an image of handwritten text. Traditional approaches contain a first segmentation
step, followed by a transcription step. Unlike printed texts, images of handwriting 
are diffucult to segment into characters. Early methods tried to compute segmentation
hypotheses for characters, for example by performing an heuristic over-segmentation,
followed by a scoring of groups of segments (e.g. in~\cite{bengio1995lerec,knerr1998hidden}).
In the nineties, this kind of approach was progressively replaced by segmentation-free 
methods, where a whole word image is fed to a system providing a sequence of scores.
A lexicon constrains a decoding step, allowing to retrieve the character sequence.
Some examples are the sliding window approach~\cite{kaltenmeier1993sophisticated}, in which features are 
extracted from vertical frames of the line image, or space-displacement neural networks
\cite{bengio1995lerec}. In the last decade, word segmentations were abandoned in favor of complete text 
line recognition with statistical language models~\cite{bunke2004offline}.

Nowadays, the standard handwriting recognition systems are multi-dimensional long short-term memory
recurrent neural networks (MDLSTM-RNNs~\cite{Graves_Schmidhuber2008}), which 
consider the whole image, alternating MDLSTM layers and convolutional layers.
The transformation of the 2D structure into a sequence is computed by a simple 
collapse layer summing the activations along the vertical axis.
The further conversion of a sequence of $T$ predictions into a sequence 
of $N \leq T$ characters is achieved by a simple mapping, involving a non-character label,
allowing to consider all possible character segmentations during training with
the connectionist temporal classification loss (CTC~\cite{Graves2006a}).
These models have become very popular and won the recent evaluations of 
handwriting recognition~\cite{maurdor,htrts,openhart}.

However, current models still need segmented text lines, and full document processing 
pipelines should include automatic line segmentation algorithms.
Although the segmentation of documents into lines is assumed in most descriptions 
of handwriting recognition systems, several papers or surveys state that it is
 a crucial step for handwriting text recognition 
systems~\cite{bosch2012statistical,likforman2007text,razak2008off}.
The need of line segmentation to train the recognition system has also motivated 
several efforts to map a paragraph-level or page-level transcript to line positions 
in the image (e.g. recently~\cite{Bluche2014b,gatos2014ground}).

In this paper, we pursue the traditional tendency to relax hard segmentation 
hypotheses in handwriting recognition systems -- from character, then word segmentation 
to full text lines -- which consistently improved the performance.
We propose a model for multi-line recognition based on the popular MDLSTM-RNNs, 
 augmented with an attention mechanism inspired from the recent models for
machine translation~\cite{bahdanau2014neural}, 
image caption generation~\cite{cho2015describing,xu2015show}, 
or speech recognition~\cite{chan2015listen,chorowski2014end,chorowski2015attention}.
In the proposed model, the \textit{``collapse''} layer is modified with an attention
network, providing weights to modulate the importance given at different positions in 
the input. 
By iteratively applying this layer to a paragraph image, the network can transcribe
each text line in turn, enabling a purely segmentation-free recognition of full 
paragraphs.

We carried out experiments on two public datasets of handwritten paragraphs: Rimes and IAM.
We report results that are competitive with the state-of-the-art systems, which
use the ground-truth line segmentation. The remaining of this paper is 
organized as follows. \sect{related} presents methods related to the one presented here,
in terms of the tackled problem and modeling choices. In \sect{mdlstm},
we introduce the baseline model: MDLSTM-RNNs. We expose in \sect{wcollapse} the 
proposed modification, and we give the details of the system. Experimental results
are reported in \sect{expes}, and followed by a short discussion in \sect{discuss},
in which we explain how the system could be improved, and present the challenge of
generalizing it to complete documents.

\section{Related Work}
\label{sec:related}

Our work is clearly related to MDLSTM-RNNs~\cite{Graves_Schmidhuber2008}, which 
we improve by replacing the simple collapse layer by a more elaborated mechanism,
itself made of MDLSTM layers. The model we propose iteratively performs an implicit
line segmentation at the level of intermediate representations.

Classical text line segmentation algorithms are mostly based on image processing techniques and heuristics
\cite{Louloudis2009b,Nicolaou2009,Papavassiliou2010,rabaev2013text,Saabni2011,Shi2009a,Zahour2007}.
However, some methods were devised using statistical models and machine learing techniques
such as hidden Markov models~\cite{bosch2012statistical}, conditional random fields~\cite{hebert2011continuous}, 
or neural networks~\cite{delakis2008text,jung2001neural,Moysset2015b,Moysset2015a}.
In our model, the line segmentation is performed implicitely and integrated in the neural network.
The intermediate features are shared by the transcription and the segmentation
models, and they are jointly trained to minimize the transcription error.

In the field of computer vision, and particularly object detection and recognition,
many neural architectures were proposed to both locate and recognize the objects,
such as OverFeat~\cite{sermanet2013overfeat} or spatial transformer networks~\cite{jaderberg2015spatial}. 
Although systems are now able to detect multiple similar objects in a scene, most methods
localize only one object, or several objects that are different. 
For scene text recognition, which is maybe the topic in computer vision closest 
to our problem, most systems still rely on a two-step process
(localization, then recognition)~\cite{ye2015text}, even though some 
approaches jointly optimize character segmentation and word 
recognition~\cite{Chen2015,wang2011end,weinman2014toward}.

Recently, many ``attention-based'' models were proposed to iteratively select 
in an encoded signal the relevant parts to make the next prediction. 
This paradigm, already suggested by Fukushima in 1987~\cite{fukushima1987neural}, 
was successfully applied to various problems such as machine translation~\cite{bahdanau2014neural}, 
image caption generation~\cite{cho2015describing,xu2015show}, 
 speech recognition~\cite{chan2015listen,chorowski2014end,chorowski2015attention},
or cropped words in scene text~\cite{lee2016recursive}. In those works, the 
localization is implicitely performed inside the neural network. 

Other papers present similar methods to read short sequence of characters (mainly digits)
with different implementations of the attention, e.g. DRAW~\cite{gregor2015draw},
RAM~\cite{ba2014multiple}, or recurrent spatial transformer networks~\cite{sonderby2015recurrent}.
We recently proposed an attention-based model to transcribe full paragraphs
of handwritten text, which predicts each character in turn~\cite{blucheSAR}.

Outputing one token at a time turns out to be prohibitive in terms of memory and 
time consumption for full paragraphs, which typically contain about 500 characters.
In the proposed system, the encoded image is not summarized as a single vector
at each timestep, but as a sequence of vectors representing full text lines.
It represents a huge speedup factor, and a comeback to the original MDLSTM-RNN
architecture, in which the collapse layer is augmented with an MDLSTM attention 
network similar to the one presented in~\cite{blucheSAR}.

\section{Handwriting Recognition with MDLSTM and CTC}
\label{sec:mdlstm}

\begin{figure}[ht]
\begin{center}
\includegraphics[width=\linewidth]{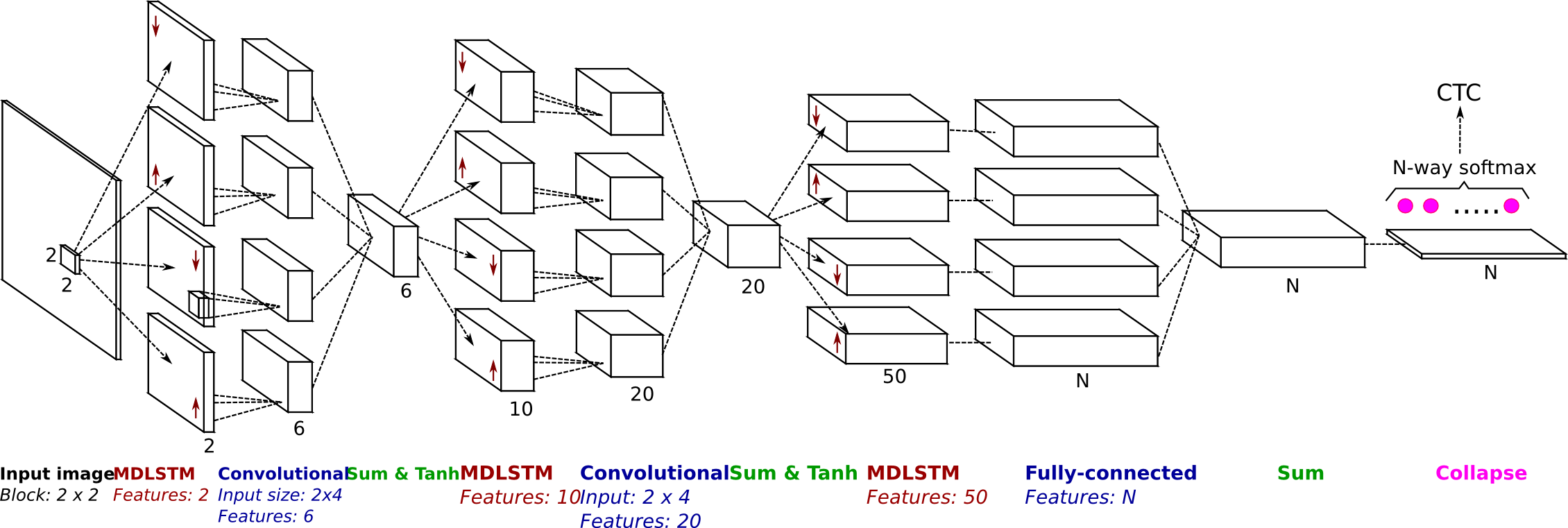}
\end{center}
\caption{MDLSTM-RNN achitecture for handwriting recognition. LSTM layers in 
         four scanning directions are followed by convolutions. 
         The feature maps of the top layer are are summed in the vertical dimension, and character
         predictions are obtained after a softmax normalization (figure from~\cite{pham2014dropout}).
         \label{fig:mdlstm}}
\end{figure}

In this section, we briefly present the MDLSTM-RNNs~\cite{Graves_Schmidhuber2008}.
MDLSTM layers generalize LSTMs to two-dimensional inputs. 
They were first introduced in the context of handwriting recognition. 
The general architecture is displayed in \fig{mdlstm}. 

The MDLSTM layers scan the input in the four possible directions.
 The LSTM cell inner state and output are computed from 
the states and outputs of previous positions in the horizontal and vertical directions.
Each LSTM layer is followed by a convolutional layer. The resolution of the learnt
representations is decreased by setting a step size of the convolutional filters 
greater than one. As the size of the feature maps decreases, the number of extracted
features increases. At the top of this network, there is one feature map 
for each character. A collapse layer sums the features along the vertical axis,
yielding a sequence of prediction vectors, normalized with a softmax activation.

In order to transform the sequence of $T$ predictions into a sequence of $N$
labels, an additionnal \textit{non-character} label is introduced,
and a simple mapping is defined to retrieve the transcription.
The connectionist temporal classification objective (CTC~\cite{Graves2006a}), 
which considers all possible labellings of the sequence, may be applied to train
the network to recognize text lines.

The 2D to 1D conversion happens in the collapsing layer, which applies a simple 
aggregation of the feature maps into vector sequences, i.e. maps of height 1.
This is achieved by a simple sum across the vertical dimension:
\begin{equation}
 \label{eqn:collapse}
 z_i = \sum_{j=1}^H a_{ij}
\end{equation}
where $z_i$ is the $i$-th output vector and $a_{ij}$ is the input feature vector at
coordinates $(i,j)$.
All the information in the vertical dimension is reduced to a single vector, 
regardless of its position in the feature maps, preventing the recognition of
multiple lines within this framwork.

\section{An Iterative Weighted Collapse for End-to-End Handwriting Recognition}
\label{sec:wcollapse}

\begin{figure}[ht]
\begin{center}
\includegraphics[width=\linewidth]{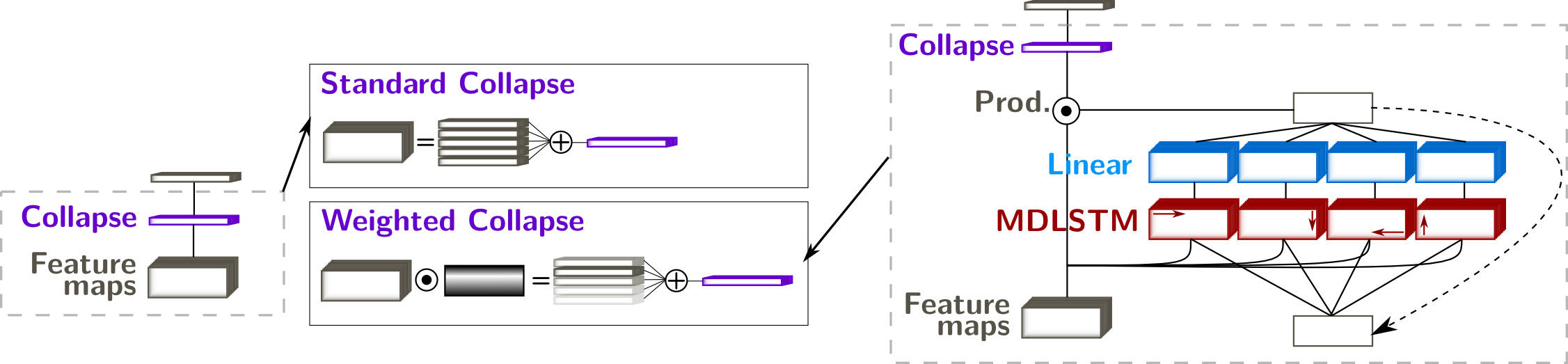}
\end{center}
\caption{Proposed modification of the collapse layer. While the standard collapse
(left, top) computes a simple sum, the weighted collapse (right, bottom) includes 
a neural network to predict the weights of a weighted sum.\label{fig:archi}}
\end{figure}

In this paper, we replace the sum of \eqn{collapse} by a weighted sum, 
in order to focus on a specific part of the input. 
The weighted collapse is defined as follows:
\begin{equation}
 \label{eqn:abcollapse}
 \ijt{z}{i}{t} = \sum_{j=1}^H \ijt{\omega}{ij}{t} a_{ij}
\end{equation}
where $\ijt{\omega}{ij}{t}$ are scalar weights between $0$ and $1$, computed 
at every time $t$ for each position $(i,j)$.
The weights are computed by a recurrent neural network, illustrated in \fig{archi},
enabling the recognition of a text line at each timestep.

This collapse, weighted with a neural network, may be interpreted as the ``attention'' 
module of an attention-based neural network similar to those 
of~\cite{bahdanau2014neural,cho2015describing,xu2015show}.
This mechanism is differentiable and can be trained with backpropagation.

Both this new architecture and the previous one are composed of an encoder 
(the MDLSTM network), an aggregation layer, and a decoder, described below.

\subsection{Encoder}

The bottom part of the architecture presented in~\sect{mdlstm} remains the same.
We can see the MDLSTM network as a feature extraction module, or encoder of the 
input image $\mathcal{I}$ into high-level features:
\begin{equation}
 \mathbf{a} = (a_{ij})_{(i,j) \in [1,W] \times [1,H]} =  Encoder( \mathcal{I} )
\end{equation}
where $(i,j)$ are coordinates in the feature maps.

In~\sect{mdlstm} a simple sum of $\mathbf{a}$ is computd by a collapse layer.
Here, we apply an attention mechanism to read text lines. 

\subsection{Attention}
\label{sec:attn_module}

The weighted collapse is an attention mechanism providing a view of the encoded image at each timestep
in the form of a weighted sum of feature vector sequences. The attention network computes a 
score for the feature vectors at every position:
\begin{equation}
\label{eqn:attention}
 \ijt{\alpha}{ij}{t} = Attention ( \mathbf{a}, \mathbf{\omega}^{(t-1)} )
\end{equation}
We refer to $\mathbf{\omega}^{(t)} = \{\ijt{\omega}{ij}{t} \}_{(1 \leq i \leq W,~1 \leq j \leq H)}$
as the attention map at time $t$, which computation depends not only on the encoded
image, but also on the previous attention features.
A softmax normalization is applied to each column:
\begin{equation}
\label{eqn:attnorm}
 \ijt{\omega}{ij}{t} = \frac{e^{\ijt{\alpha}{ij}{t}}}{\sum_{j'} e^{\ijt{\alpha}{ij'}{t}}}
\end{equation}

This module is applied several times to the features from the encoder. 
The output of the attention module at iteration $t$, computed with \eqn{abcollapse},
is a sequence of feature vectors, intended to represent a text line. 
Therefore, we may see this module as a soft line segmentation neural network.
The advantages over the neural networks trained for line 
segmentation~\cite{delakis2008text,jung2001neural,Moysset2015a,Moysset2015b}
are that \textit{(i)} it works on the same features as those used for the transcription
(multi-task encoder) and \textit{(ii)} it is trained to maximize the transcription 
accuracy (i.e. more closely related to the goal of handwriting recognition systems,
and easily interpretable).

\subsection{Decoder}

The final component of this architecture is a decoder, which predicts a character 
sequence from the feature vectors.
\begin{equation}
 \mathbf{y} = Decoder( \mathbf{z} )
\end{equation}
where $\mathbf{z}$ is the concatenation of $z^{(1)}, z^{(2)}, \ldots, z^{(T)}$.
Alternatively, the deocder may be applied to $z^{(i)}$s sub-sequences to get 
$y^{(i)}$s and $\mathbf{y}$ is the concatenation of $y^{(1)}, y^{(2)}, \ldots, y^{(T)}$.

In the standard MDLSTM architecture of \sect{mdlstm}, the decoder is a simple 
softmax. However, a Bidirectional LSTM (BLSTM) decoder could be applied to the 
collapsed representations. This is particularly interesting in the proposed
model, as the BLSTM would potentially process the whole paragraph, 
allowing a modeling of dependencies across text lines.

\subsection{Training}

\begin{figure}[ht]
\begin{center}
\includegraphics[width=\linewidth]{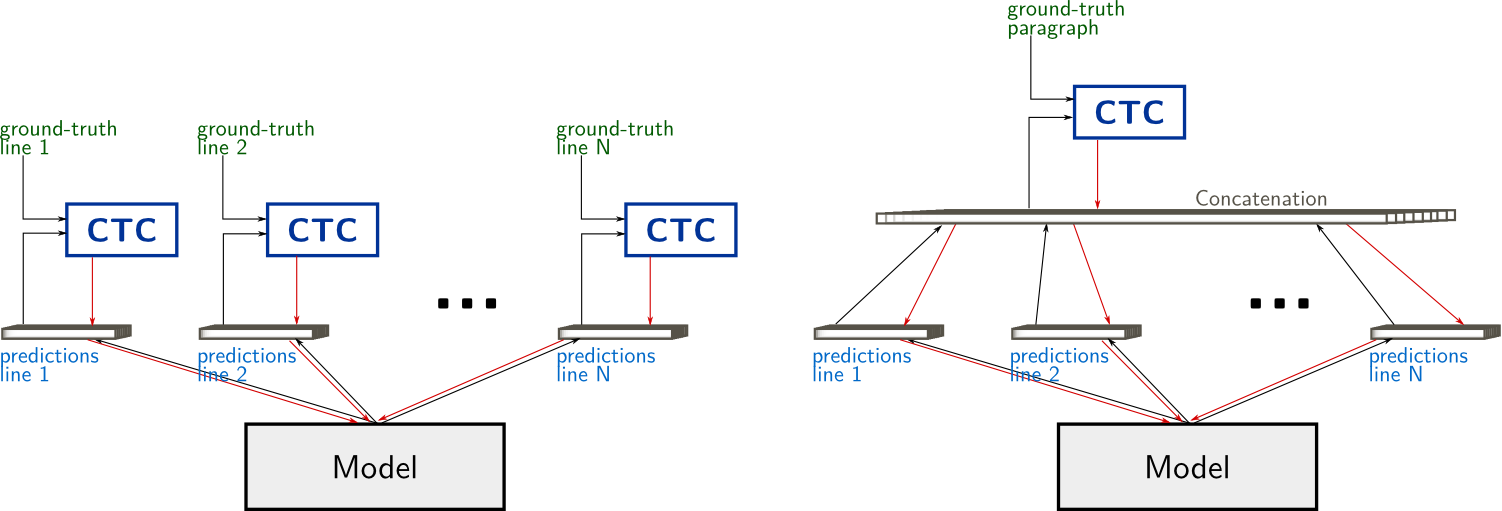}
\end{center}
\caption{Training strategies.
If a ground-truth is available at
the line level, the CTC objective function may be applied on line segments independently
(left). If the ground-truth is only available at the paragraph level, the CTC objective is 
applied to the concatenation of all line predictions (right).
\label{fig:train_strat}}
\end{figure}

This model can be trained with CTC. If the line breaks are known in the transcript,
the CTC could by applied to the segments corresponding to each line prediction,
with the line transcript. Moreover, it will enforce the prediction at each timestep
to correspond to a complete text line.
Otherwise, one can directly apply CTC to the whole paragraph.
The different training strategies of this model are illustrated in \fig{train_strat}.

In this work, we mainly investigated the second strategy, with CTC training at the 
paragraph level, and with a BLSTM decoder applied to the concatenation of all 
collapsing steps, for reasons developed in the next section.

\subsection{Limitations}
\label{sec:limits}

Compared to the model presented in~\cite{blucheSAR}, the iterative decoder requires
one step for each text line instead of one step for each character, which represents
a huge speedup of a factor 20-30. However, we loose the ability to handle arbitrary
reading orders.
Moreover, in this version, the model does not predict a \textit{``stop''} token.
Thus, the network predicts an arbitrary number of sequence $T$, fixed by the 
experimenter. In our experiments, $T$ corresponds to the maximum number of 
lines in the dataset. The BLSTM decoder is applied to these $T$ sequences and 
was efficient to ignore the supplementary lines in shorter paragraphs. 
In numerous cases, we observed that during these additional steps, the 
attention was located in interlines, were the decoder can easily only predict
non-characters. However, missing the ability to determine automatically the number 
of required steps is an important limitation, which should be fixed in future work.

Finally, the collapsing paradigm forces the model to output sequences that span the 
whole width of the image. We may replace the column-wise softmax of \eqn{attnorm}
with a sigmoid to ignore some parts of the input, for shorter lines for example,
but a refined mechanism that selects only a portion of the image will become 
crucial to handle complete documents with complex layouts. That issue will be discussed
in more details in \sect{discuss}.

\section{Experiments}
\label{sec:expes}

\subsection{Experimental Setup}

We carried out the experiments on two public databases.
The IAM database~\cite{iam} is made of handwritten English texts copied from the LOB corpus. 
There are 747 documents (6,482 lines) in the training set, 
116 documents (976 lines) in the validation set and 
336 documents (2,915 lines) in the test set.
The Rimes database~\cite{rimes} contains handwritten letters in French. 
The data consists of a training set of 1,500 paragraphs (11,333 lines), 
and a test set of 100 paragraphs (778 lines). 
We held out the last 100 paragraphs of the training set as a validation set.

The networks have the following architecture. The encoder first computes a 2x2 
tiling of the input and alternate MDLSTM layers of 4, 20 and 100 units and 
2x4 convolutions of 12 and 32 filters with no overlap. The last 
layer is a linear layer with 80 outputs for IAM and 102 for Rimes. The attention 
network is an MDLSTM network with 2x16 units in each direction followed by a 
linear layer with one output, and a softmax on columns (\eqn{attnorm}). 
The decoder is a BLSTM network with 256 units.
The networks are trained with RMSProp~\cite{tielmann} with a base learning rate of
$0.001$ and mini-batches of 8 examples, to minimize the CTC loss over entire paragraphs.

In the following, we study the impact of adding a BLSTM decoder, and an attention-based
collapse (\sect{expedecoder}), we compare our method to the baseline results on
automatic and ground-truth line segmentation (\sect{expelineseg}) and we present
a comparison of our system to the state of the art (\sect{experesult}).

\subsection{Impact of the Decoder}
\label{sec:expedecoder}

As explained in \sect{limits}, in our model, the weighted collapse method is
followed by a BLSTM decoder. In this experiment, we compare the baseline system 
(standard collapse followed by a softmax) with the proposed model. 
In order to dissociate the impact of the weighted collapse from that of the BLSTM 
decoder, we also trained an intermediate architecure with a BLSTM layer after the 
standard collapse, but still limited to text lines. 

\begin{table}[htb]
\begin{center}
\begin{tabular}{|r|l|l|c|}\hline
\textbf{Database} & \textbf{Collapse} & \textbf{Decoder} & \textbf{CER\%} \\\hline
 \textbf{IAM}   & Standard  & Softmax & 8.4 \\
                & Standard  & BLSTM + Softmax & 7.5 \\
                & Attention & BLSTM + Softmax & 6.8 \\\hline
 \textbf{Rimes} & Standard  & Softmax & 4.9 \\
                & Standard  & BLSTM + Softmax & 4.8 \\
                & Attention & BLSTM + Softmax & 2.5 \\\hline
\end{tabular}
 \end{center}
\caption{Character Error Rates (\%) of CTC-trained RNNs on 150 dpi images. The \textit{Standard}
models are trained on segmented lines. The \textit{Attention} models are trained
 on paragraphs.\label{tab:collapse}}
\end{table}

The character error rates (CER\%) on the validation sets are reported in 
\tab{collapse} for 150dpi images.
We observe that the proposed model outperforms the baseline by a large margin 
(relative 20\% improvement on IAM, 50\% on Rimes), and that the gain may be 
attributed to both the BLSTM decoder, and the attention mechanism.

\subsection{Impact of Line Segmentation}
\label{sec:expelineseg}

Our model performs an implicit line segmentation to transcribe paragraphs. 
The baseline considered in the last section is somehow cheating, because it 
was evaluated on the ground-truth line segmentation. 
In this experiment, we add to the comparison the baseline models evaluated in
a real scenario where they are applied to the result of an automatic line 
segmentation algorithm.

\begin{table}[htb]
\begin{center}
\begin{tabular}{|r|r||c||c|c|c||c|}
\hhline{-:-::-::-:-:-::-:} 
 \textbf{Database} & \textbf{Resolution} & \textbf{GroundTruth} & \textbf{Projection} & \textbf{Shredding} & \textbf{Energy} & \textbf{This work} \\
\hhline{=:=::=::=:=:=::=:}  
 \textbf{IAM}    & 150 dpi &  8.4 & 15.5 &  9.3 & 10.2 & 6.8 \\ 
                 & 300 dpi &  6.6 & 13.8 &  7.5 &  7.9 & 4.9 \\
\hhline{=:=::=::=:=:=::=:}
  \textbf{Rimes} & 150 dpi &  4.8 &  6.3 &  5.9 &  8.2 & 2.8 \\ 
                 & 300 dpi &  3.6 &  5.0 &  4.5 &  6.6 & 2.5 \\\hline 
 \end{tabular}
 \end{center}
\caption{Character Error Rates (\%) of CTC-trained RNNs on ground-truth lines and 
 automatic segmentation of paragraphs with different resolutions.
The last column contains the error rate of the attention-based model presented in this 
work, without an explicit line segmentation.\label{tab:autoseg}}
\end{table}

In \tab{autoseg}, we report the CERs obtained with the ground-truth line positions,
with three different segmentation algorithms, and with our end-to-end system,
on the validation sets of both databases with different input resolutions. 
We see that applying the baseline networks on automatic segmentations
increases the error rates, by an absolute 1\% in the best case. We also observe 
that the models are better with higher resolutions. 

\begin{figure}[ht]
\begin{center}
\includegraphics[width=\linewidth]{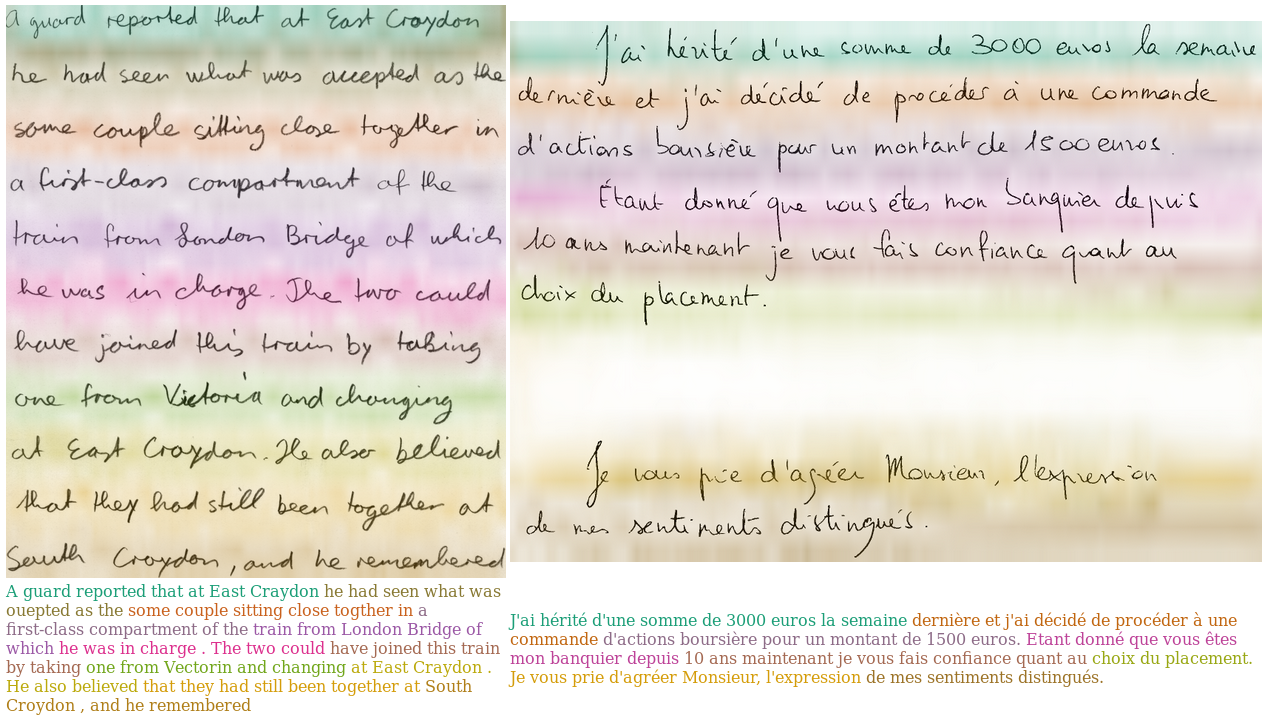} 
\caption{Transcription of full paragraphs of text and implicit line segmentation 
learnt by the network on IAM (left) and Rimes (right). Best viewed in color.\label{fig:para}}
 \end{center}
\end{figure}

Our models yield better performance than methods based on an explicit and 
automatic line segmentation, and comparable or better results than with 
ground-truth segmentation, even with a resolution divided by two. 
In \fig{para}, we display a visualisation of the implicit line 
segmentation computed by the network. Each color corresponds to one step of the 
iterative weighted collapse. On the images, the color represents the weights 
given by the attention network (the transparency encodes their intensity). 
The texts below are the predicted transcriptions, and chunks are colored according
to the corresponding timestep of the attention mechanism.

\subsection{Comparison to Published Results}
\label{sec:experesult}

In this section, we also compute the word error rates (WER\%) and evaluate our 
models on the test sets in order to compare the proposed approach to existing 
systems. For IAM, we applied a $3$-gram language model with a lexicon of 50,000 words, 
trained on the LOB, Brown and Wellington corpora\footnote{ The parts of the LOB corpus 
used in the validation and evaluation sets were removed.}. This language 
model has a perplexity of 298 and OOV rate of 4.3\% on the validation set 
(329 and 3.7\% on the test set).

The results are presented in \tab{finalresultsrimes} for Rimes and in 
\tab{finalresultsiam} for IAM, for different input resolutions. When comparing
the error rates, it is important to note that all systems in the litterature
used an explicit (ground-truth) line segmentation and a language model. 
\cite{doetschfast,Kozielski2013a,MessinaOOV2014} used an hybrid
character/word language model to tackle the issue of out-of-vocabulary words.
Moreover, all systems except \cite{MessinaOOV2014,pham2014dropout} carefully 
pre-processed the line image (e.g. corrected the slant or skew, normalized the height, ...),
whereas we just normalized the pixel values to zero mean and unit variance.
Finally, \cite{phdthesis} is a combination of four systems.

\begin{table}[!htb]
 \begin{center}
\footnotesize
\begin{tabular}{|l r||c|c||c|c|}
\hhline{~~|-|-||-|-|}
\multicolumn{2}{r|}{} & \multicolumn{2}{c||}{\textbf{Validation}} & \multicolumn{2}{c|}{\textbf{Test}} \\
\multicolumn{2}{r|}{} & \textbf{WER\%} & \textbf{CER\%} & \textbf{WER\%} & \textbf{CER\%} \\
\hhline{--::=:=::=:=:}      
\textbf{150 dpi} & no language model & 12.7 & 2.8 & 13.6 & 3.2 \\
\hhline{==::=:=::=:=:}
\textbf{300 dpi} & no language model & 12.0 & 2.5 & 12.6 & \textbf{2.9} \\
\hhline{==::=:=::=:=:}
& Bluche, 2015 \cite{phdthesis}       & 11.2 & 3.3 & \textbf{11.2} & 3.5   \\
& Pham et al., 2014 \cite{pham2014dropout} & - & - & 12.3 & 3.3 \\
& Doetsch et al., 2014 \cite{doetschfast}     & - & - & 12.9 & 4.3 \\
& Messina \& Kermorvant, 2014 \cite{MessinaOOV2014}  & - & - & 13.3 & - \\
& Kozielski et al. 2013 \cite{Kozielski2013a}  & - & - & 13.7 & 4.6 \\\hline
\end{tabular}
\end{center}
\caption{Final results on Rimes database}
\label{tab:finalresultsrimes}
\end{table} 

On Rimes, the system applied to 150 dpi images already outperforms the state of 
the art in CER\%, while being competitive in terms of WER\%. The system for 300 dpi 
images is comparable to the best single system~\cite{pham2014dropout} in WER\% 
with a significantly better CER\%.

\begin{table}[!htb]
 \begin{center}
\footnotesize
\begin{tabular}{|l r||c|c||c|c|}
\hhline{~~|-|-||-|-|}
\multicolumn{2}{r|}{} & \multicolumn{2}{c||}{\textbf{Validation}} & \multicolumn{2}{c|}{\textbf{Test}} \\
\multicolumn{2}{r|}{} & \textbf{WER\%} & \textbf{CER\%} & \textbf{WER\%} & \textbf{CER\%} \\
\hhline{--::=:=::=:=:}
\textbf{150 dpi} &  no language model    & 22.4 & 6.8 & 29.5 & 10.1 \\
                 &  with language model  & 13.8 & 4.7 & 16.6 &  6.5 \\
\hhline{==::=:=::=:=:}
\textbf{300 dpi} &  no language model    & 17.7 & 4.9 & 24.6 &  7.9 \\
                 &  with language model  & 13.1 & 3.5 & 16.4 &  5.5 \\
\hhline{==::=:=::=:=:}
& Bluche, 2015 \cite{phdthesis}               &  9.6 & 3.3 & \textbf{10.9} & \textbf{4.4}   \\
& Doetsch et al., 2014 \cite{doetschfast}             & \textbf{8.4} & \textbf{2.5} & 12.2 & 4.7 \\
& Kozielski et al. 2013 \cite{Kozielski2013a}          &  9.5 & 2.7 & 13.3 & 5.1 \\
& Pham et al., 2014 \cite{pham2014dropout}         & 11.2 & 3.7 & 13.6 & 5.1 \\
& Messina \& Kermorvant, 2014 \cite{MessinaOOV2014}          &  -   &  -  & 19.1 & - \\
& Espana-Boquera et al., 2011 \cite{espana2011improving}     & 19.0 &  -  & 22.4 & 9.8 \\\hline
\end{tabular}
\end{center}
\caption{Final results on IAM database\label{tab:finalresultsiam}}
\end{table} 

On IAM, the language model turned out to be quite important, probably because 
there is more variability in the language\footnote{ A simple language model yields
a perplexity of 18 on Rimes~\cite{phdthesis}.}. On 150 dpi images, the results are not too far from 
the state of the art results. The WER\% does not improve much on 300 dpi images, 
but we get a lower CER\%. When analysing the errors, we noticed that there is a lot 
of punctuation in IAM, which was often missed by the attention mechanism.

\section{Discussion}
\label{sec:discuss}

As already discussed in \sect{limits}, the proposed model can transcribe complete
paragraphs without segmentation and is orders of magnitude faster that the 
model of \cite{blucheSAR}. However, the mechanism cannot handle arbitrary
reading orders. Rather, it implements a sort of implicit line segmentation.
In the current implementation, the iterative collapse runs for a fixed number of 
timesteps. Yet, the model can handle a variable number of text lines, and, interestingly,
the focus is put on interlines in the additional steps. 
A more elegent solution should include the prediction of a binary variable indicating
when to stop reading.

Our method was applied to paragraph images, so a document layout analysis should
be applied to detect those paragraphs before applying the model. 
Naturally, the next step should be the transcription of complex documents without
an explicit or assumed paragraph extraction. 
The limitation to paragraphs is inherent to this system.
Indeed, the weighted collapse always outputs sequences corresponding to the whole width 
of the encoded image, which, in paragraphs, may correspond to text lines.
In order to switch to full documents, several issues arise.

First, the size of the lines are determined by the size of the text block. 
Thus a method should be devised to only select a smaller part of the feature maps,
representing only the considered text line.
This is not possible in the presented framework. A potential solution could come 
from spatial transformer networks \cite{jaderberg2015spatial}, performing a differentiable
crop. However, that method is based on learning a grid transformation, with a 
fixed grid size, while we would like to crop variable-sized parts.
Another solution would be hierarchical and comprise a first attention at the 
text block level, and a second one at the line level inside the block. Note that 
we would probably still need to crop the text block. 
In a different direction, we could also abandon the differentiability requirement,
and learn to predict the crops with reinforcement learning techniques.

On the other hand, training will in practice become more difficult, not only because of
the complexity of the task, but also because the reading order in complex documents 
cannot be exactly inferred in many cases. Even defining arbitrary rules can be 
tricky. Therefore, the matching of predictions with ground-truth texts should 
be addressed. 

Finally, we would like to point out some important factors to take into account 
when training the presented model. Because CTC training may have difficulties to
find good alignments and to have the network predict actual characters and not 
only non-character symbols, the convergence is much faster with a pre-trained encoder. 
For example, one can first train an MDLSTM-RNN with the standard collapse on text lines, 
and finetune it with the attention-based collapse on paragraphs in a second step. 
However, training the attention model on full paragraphs directly was actually
not easy, and we found curriculum methods useful. Before switching to full paragraphs,
we had to train for a few epochs on two or three lines to initiate the attention 
mechanism. This should also be taken into account for complete documents. A good 
curriculum will be harder to design, and probably crucial. Nonetheless, the 
amount of data used in our experiments is quite limited, and careful training 
might become less important with more data.

\section{Conclusion}

We have presented a model to transcribe full paragraphs of handwritten texts 
without an explicit line segmentation. Contrary to classical methods relying on
a two-step process (segment then recognize), our system directly considers the 
paragraph image without an elaborated pre-processing, and outputs the complete 
transcription. We proposed a simple modification of the collapse layer in the 
standard MDLSTM architecture to iteratively focus on single text lines. 
This implicit line segmentation is learnt with backpropagation along with the 
rest of the network to minimize the CTC error at the paragraph level. 

We reported comparable error rates to the state of the art on two public databases.
After switching from explicit to implicit character, then word segmentation for 
handwriting recognition, we showed that line segmentation can also be learnt inside
the transcription model. The next step towards end-to-end handwriting recognition 
is now at the full page level.

\bibliographystyle{plain}
\small
\bibliography{ab_collapse}

\end{document}